# Responsible AI: Portraits with Intelligent Bibliometrics

Yi Zhang, *Senior Member, IEEE*, Mengjia Wu, Guangquan Zhang, Jie Lu, *Fellow, IEEE*

***Abstract*—Shifting the focus from principles to practical implementation, responsible artificial intelligence (AI) has garnered considerable attention across academia, industry, and society at large. Despite being in its nascent stages, this emerging field grapples with nebulous concepts and intricate knowledge frameworks. By analyzing three prevailing concepts - explainable AI, trustworthy AI, and ethical AI, this study defined responsible AI and identified its core principles. Methodologically, this study successfully demonstrated the implementation of leveraging AI's capabilities into bibliometrics for enhanced knowledge discovery and the cross-validation of experimentally examined models with domain insights. Empirically, this study investigated 17,799 research articles contributed by the AI community since 2015. This involves recognizing key technological players and their relationships, unveiling the topical landscape and hierarchy of responsible AI, charting its evolution, and elucidating the interplay between the responsibility principles and primary AI techniques. An analysis of a core cohort comprising 380 articles from multiple disciplines captures the most recent advancements in responsible AI. As one of the pioneering bibliometric studies dedicated to exploring responsible AI, this study will provide comprehensive macro-level insights, enhancing the understanding of responsible AI while furnishing valuable knowledge support for AI regulation and governance initiatives.**

***Impact Statement* — This study stands as one of the pioneering investigations into responsible artificial intelligence (AI), utilizing large-scale and comprehensive bibliometric studies. It develops an analytical framework that elaborates cutting-edge intelligent bibliometric models. The study analyzes responsible AI from diverse aspects, including the recognition of key technological players, the identification of topical landscape, hierarchy, and evolutionary pathways, and the profile of a core cohort about responsible AI. Of particular significance is the empirical insights of the bibliometric portraits for responsible AI, which would benefit the broader AI community and relevant stakeholders to understand this emerging topic at a macro level and provide essential knowledge support for AI regulation and governance initiatives.**

## I. Introduction

ARTIFICIAL intelligence (AI), with its incredible capabilities in learning patterns from large-scale data and training human-like intelligent models, is revolutionizing the human society from automation to digitalization with autonomous and self-awareness machines. The unprecedented success of generative AI, particular, ChatGPT, further triggers a global debate on AI and its profound societal impact, such as risks and threats to the broad society [1], and what to regulate and how [2].

Since the early beginning of the AI boom in the 2010s, in fact, global government agencies, giant IT companies, research institutions, professional societies, and non-profit organizations have started to produce a series of principles and guidelines for *AI ethics* [3], broadly touching the good of humanity, individuals, societies, and the environment[1]. While AI ethics provides a visionary blueprint to describe what are the right things that we expect AI to do, the society urged an executive and tactical plan for how we can do.

This urgent need fosters the rise of *responsible AI* – a plan for both AI systems and their practitioners. This responsibility entails an AI system's fundamental functionality that enables AI to reason about and act according to human values. It further underscores the accountability and awareness of AI researchers and practitioners on the direct societal impact of AI [4]. Emphasizing a pragmatic and application-driven approach, the industry simplifies responsible AI as approaches to designing, developing, and deploying AI under a legal, ethical, and trustworthy framework [5], and has produced consultation reports and industry standards to support the implementation of responsible AI practices [6, 7].

The AI community's response to responsible AI appears to adopt a spontaneous bottom-up paradigm. There have been numerous algorithms and models, placing emphasis on their responsibility-related functionalities, e.g., the protection of privacy-sensitive data [8] and the promotion of fair decision support [9]. From a technical standpoint, impressive endeavors on conceptualizing AI's social responsibility [10] and

This paper was submitted on January 15, 2024.

This work was supported by the Commonwealth Scientific and Industrial Research Organization (CSIRO), Australia, in conjunction with the National Science Foundation (NSF) of the United States, under CSIRO-NSF #2303037; and the Australian Research Council (ARC) under Laureate Project FL190100149.

Y. Z., M. W., G. Z., and J. L. are with the Australian Artificial Intelligence Institute (AAII), University of Technology Sydney (UTS), Sydney, Australia. (emails: Yi.Zhang@uts.edu.au, Mengjia.Wu@uts.edu.au, Guangquan.Zhang@uts.edu.au, Jie.Lu@uts.edu.au).

[1] The United Nations Educational, Scientific and Cultural Organization (UNESCO) recommended four core values of AI systems: Human rights and human dignity, living in peaceful, ensuring diversity and inclusiveness, and environment and ecosystem flourishing. See the website: https://www.unesco.org/en/artificial-intelligence/recommendation-ethics

informing policy development for responsible AI [11] have created prominent impact. However, in this rapidly developing and emerging area, the AI community has not reached a consensus regarding the definitions, concepts, and technologies pertaining to responsible AI, despite some roadmaps outlining its application in specific sectors, e.g., agriculture [12] and healthcare [13]. Consequently, an investigation into the topical landscape, hierarchy, and evolution of responsible AI, along with an exploration of its key technological contributors, will bring fresh and expert insights to the development of a comprehensive strategy for AI governance and regulation.

This study aims to fulfill the gap by developing an intelligent bibliometrics-based analytical framework, designed to unveil comprehensive portraits of responsible AI, and discover empirical insights. Intelligent bibliometrics refers to the development and applications of AI-empowered computational models for analyzing large-scale bibliometric data such as scientific articles and patent documents [14]. Notable models include scientific evolutionary pathways, enabling the trace of technological change through streaming data analytics [15], topical hierarchical trees, employed to profile the knowledge structure of a research domain using density-based hierarchical community detection [16], and knowledge trajectory prediction using graph analytics [17]. Intelligent bibliometrics has achieved significant success in broad science, technology, and innovation studies, e.g., tracking topical disruption, evolution, and resilience of coronavirus research [18] and unveiling digital transformation enabling capabilities [19].

This study created a comprehensive set of responsible AI-related portraits, with the objectives of recognizing key players and their relationships, identifying the topical landscape and hierarchy of responsible AI, charting its evolutionary pathways, and elucidating the interplay between responsibility principles and primary AI techniques. We combined research articles collected from Scopus and the Web of Science Core Collection and cross-referenced with the widely acknowledged academic knowledge graph OpenAlex. This yielded a dataset comprising 17,799 distinct articles on responsible AI.

Through this investigation, we drew the technical backbones of responsible AI, spanning from data architectures (data science) to analytical algorithms (machine learning), and from fundamental theories (mathematics) to system applications (computer networks). With a strong focus on privacy and security studies, we observed a dominance of contributions from China, the United States, and India, collectively accounting for over 50% of total publications. Our findings also highlighted concerted efforts in addressing the issues of explainability, transparency, trustworthiness, and reliability, particularly emanating from the machine learning community. Furthermore, our focus on a core cohort of 380 articles explicitly referencing the term "responsible AI" unveiled the cross-disciplinary nature of the field. From the perspective of a complex eco-system, the chaotic connections among terms from computer science and the humanities and social science disciplines may underscore the gradual establishment of responsible AI as a new knowledge area. Top-tier universities and prominent IT companies from the West are playing pivotal roles in driving its development.

This study stands as one of the pioneering investigations into responsible AI, utilizing large-scale bibliometric analysis. It develops an analytical framework that elaborates cutting-edge intelligent bibliometric models. The study encompasses diverse aspects, including the recognition of key technological players, and the identification of topical landscape, hierarchy, and evolutionary pathways. Of particular significance is the empirical insights of the bibliometric portraits for responsible AI. These valuable insights would benefit the broader AI community and relevant stakeholders to understand this emerging topic at a macro level and provide essential knowledge support for AI regulation and governance initiatives.

This paper is organized as follows: We conceptualized responsible AI with definitions, principles, and practices in Section II. Section III introduces the data source and outlines the methodologies of intelligent bibliometrics. Section IV presents the results of our study, drawing a series of portraits on responsible AI, regarding its key players and topical landscape, hierarchy, and evolutionary pathways. We concluded this study, discussed technical and political implications, and anticipated future directions in Section V.

## II. Conceptualization: Responsible AI

The pervasive adoption of AI techniques across public and private sectors has accelerated the interactions between AI and human activities, thereby amplifying the ethical dimensions of AI, e.g., imbalanced training data affecting underrepresented population groups [20] and the exposure and preservation of privacy-sensitive data [21]. Consequently, there has been a growing call for advanced algorithmic development and upgrades to proactively address these widely discussed ethical concerns, and it has garnered positive and constructive responses from the AI community.

This section is to conceptualize the definition and principles of responsible AI by distinguishing it from related concepts such as explainable AI, trustworthy AI, and ethical AI and theoretically elaborating academic debates, industry standards, and domain practices.

### *A. Responsible AI: Conceptual Analysis*

Before the advent of responsible AI, the AI community has shown heightened interest in addressing societal concerns and creating positive societal impact through the rise of several conceptual trends. This section is to analyze three prominent concepts: explainable AI, trustworthy AI, and ethical AI.

*Explainable AI* highlights an AI system's capacity to be comprehensible to humans, not only in explaining its decisions, recommendations, and predictions, but also in delineating the process of generating these actions [22]. In the AI community, there has been a concerted effort to devise novel approaches to understand feature sensitivity and influence [23] and navigate the trade-off between model performance and explainability [24]. Often referred to by alternative names such as transparent and accountable AI [25], explainability is frequently intertwined with reliability and trust in the literature [26].

*Trustworthy AI* emphasizes the safety, security, fairness, and privacy protection of an AI system, and underscores the crucial role of institutional, software, and hardware mechanisms in constructing a reliable and accessible AI ecosystem [27]. The pursuit of trustworthy AI aligns with a shared interest in detecting and preventing attacks and protecting data privacy [8]. This concept has far-reaching influence in cybersecurity, internet of things, social media, etc. [28]

*AI ethics* (and ethical AI) transcends the boundaries of the AI community and marks AI's initiative societal interactions, with the intensive involvement of academia, industry, government, and the public. Unlike a purely technical initiative, AI ethics encapsulates the collective understanding and societal expectation regarding AI and its applications [29]. Some of its core ethical principles include: transparency, justice & fairness, non-maleficence, responsibility, privacy, beneficence, freedom & autonomy, trust, sustainability, dignity, and solidarity [3]. The comprehensive set of ethical is in line with fundamental human values. However, some ongoing debates within the community argue that "we should not yet celebrate consensus around high-level principles that hide deep political and normative disagreement" [30].

### B. Responsible AI: Principles and Practices

Responsible AI addresses this argument by transitioning from principles to practices, viewing AI as a socio-technical ecosystem encompassing a comprehensive interplay of techniques, developers, and practitioners [4, 31]. Aligned with the values inherent in explainable AI, trustworthy AI, and ethical AI, this socio-technical ecosystem requires diversity and inclusiveness to foster cross-boundary innovation. Thus, responsible AI goes beyond merely "ticking ethical boxes" or incorporating add-on features in AI systems [31].

Conceptually, the practice of responsible AI follows the management architecture of an AI system, traversing ethical guidelines, risk controls, data governance, and training and education [32]. Notably, Microsoft [6] defined six principles of responsible AI standards: accountability, transparency, fairness, reliability & safety, privacy & security, and inclusiveness. These principles are accompanied by detailed targets specifying the stakeholders to whom each standard applies, executive requirements, and stepwise instructions with milestones. Similarly, addressing specific pain points faced by practitioners, Accenture proposed four pillars for responsible AI implementations: organizational, operational, technical, and reputational [33].

In academia, Cheng, et al. [10] outlined the root causes of AI's responsible issues, such as data bias (e.g., inconsistency, lack of transparency, and imbalance), over-simplified loss functions, inappropriate evaluation metrics, and mis-interpretation. Despite the absence of a unanimous consensus, the AI community has undertaken substantial efforts to individually address some of these issues, e.g., adversarial learning to foster fair representation by mitigating biases in features [34], deep uncertainty quantification [35], and self-explainable graph neural networks [36].

In this study, we adhered to the definition and interpretation of responsible AI put forth by Dignum [31]. However, we augmented this perspective by integrating the Microsoft responsible AI standards [6] and the conceptualization of socially responsible AI [10]. This approach enables us to encompass the viewpoints from both industry and academia. We, thus, defined:

- **Responsible AI** *is a socio-technical ecosystem with AI techniques, developers, and practitioners. The system enables AI with fundamental functionalities to reason about and act according to human values, and fosters accountability and awareness among AI developers and practitioners regarding AI's societal impact.*
- **The responsibility principles** *include accountability, explainability, transparency, fairness, intelligibility, un-bias, non-discrimination, reliability, safety, privacy, security, inclusiveness, and accessibility.*

## III. Data and Methodology

This study is to investigate the AI community's efforts on responsible AI, including proposing principles and standards, developing responsible models, and implementing applications to tackle real-world responsibility-related concerns. To achieve this, we applied a search strategy for collecting and cross-referencing relevant research articles and developed an intelligent bibliometrics-based analytical framework to analyze the data and generate a comprehensive set of portraits on responsible AI.

### A. Data

Recognized as two primary sources of scientific databases, the *Web of Science* (WoS) Core Collection[2] and *Scopus*[3] offer a comprehensive coverage of journal articles and conference papers, each with distinct emphases. The WoS Core Collection highlights curating high-quality journals indexed by the Science Citation Index and Social Science Citation Index, while Scopus enables a broader scope by encompassing a diverse array of conference proceedings, a platform where the AI community actively disseminates research findings. Therefore, we decided to leverage both data sources to gain their complementary strengths. Referring to the definition of responsible AI and its principles, we built the skeleton of the search string as follows:

*responsib* OR accountab* OR explainab* OR transparan* OR fair* OR intelligib* OR bias OR discriminat* OR reliab* OR safety OR privacy OR security OR inclusive* OR accessib**

Despite the extensive history of AI, we chose to commence our search from January 1, 2015, for the following reasons:

- Deep learning triggered the AI boom around 2011, but societal concerns regarding AI ethics did not materialize until +at least 2015 [37].
- The earliest literature on responsible AI appears to emerge around 2017 [38].

[2] https://clarivate.com/products/scientific-and-academic-research/research-discovery-and-workflow-solutions/webofscience-platform/web-of-science-core-collection/

[3] https://www.scopus.com/home.uri

We conducted our search on October 24, 2023, and collected 12,952 articles from Scopus and 10,043 articles from the WoS Core Collection. Subsequently, we merged the two datasets and eliminated duplicates, resulting in a refined collection of 18,298 distinct articles. The detailed search strategy is in Table I, with the following notes:

- In contrast to social sciences and humanities, the titles of research articles in computer science are typically direct and straightforward, devoid of rhetorical devices, artistic references, or embellishments. However, abstracts may introduce extraneous information such as background details and practical implications, potentially adding noise to the search results. Thus, our search focused exclusively on titles and keywords.
- Scopus utilizes the All Science Journal Classification (ASJC) system to categories journals and conference proceedings, linking them to related articles. The WoS Core Collection features its own subject category system. While both systems contain an AI-related category, they apply distinct classification criteria for journals and conferences. This motivated our decision to utilize both databases to ensure a comprehensive coverage.
- We concentrated on research articles published in journals and conference proceedings and excluded other source types (e.g., books, book series, and reports) and document types (e.g., editorials, review articles, notes, and letters).
- We only focused on English publications.

TABLE I
SEARCH STRATEGY

| ***Constraints*** | ***Scopus*** | ***WoS Core Collection*** |
|---|---|---|
| Search fields | titles and keywords | |
| Language | English | |
| Time | 01/01/2015 – 24/10/2023 | |
| Subject area/category | computer science – artificial intelligence | computer science, artificial intelligence |
| Document type | article OR conference paper | article OR proceeding paper |
| Source type | journal OR conference proceeding | n/a |
| Raw results | 12,952 | 10,043 |
| Combined results | 18,298 | |

Academic knowledge graphs, with structured bibliographical attributes (e.g., titles, abstracts, authors, and references), disambiguated name entities (e.g., country, institution, and author names), and predefined document-topic linkages, prove highly advantageous in the pre-processing of bibliometric data. OpenAlex[4] is an open and free-access academic knowledge graph, with a comprehensive integration of the Microsoft Academic Graph (MAG; one representative academic graph produced by Microsoft), Crossref, and some other scientific data sources (e.g., PubMed). OpenAlex inherits the hierarchical topical system known as the Field of Study (FoS) from MAG. This system consists of a series of topical tags created by hierarchical topic models [39], with each article containing at least one FoS tag.

In this study, we leveraged OpenAlex APIs to match the 18,298 combined articles from Scopus and WoS, using their Digital Object Identifiers (DOIs). This process returned 17,799 articles, with the missing articles primarily attributed to the absence of DOIs and the early publications of 2024. We, then, retrieved the complete bibliographical information of the 17,799 articles and extracted key entities, including 49,792 authors, 6,068 affiliations, 137 countries and regions, and 8,523 FoS tags.

### *B. Intelligent Bibliometrics*

As an emerging cross-disciplinary research field, we defined:

- **Intelligent bibliometrics** [14] *refers to the development and applications of computational models that elaborate AI and data science techniques with bibliometric data (e.g., scientific articles and patent documents) and indicators (e.g., citations, keywords, and authorships).*

Intelligent bibliometrics primarily targets real-world science, technology, and innovation (ST&I) challenges, and it includes diverse methodologies, such as embedding-based ST&I topic extraction [40], network analytics and graph learning for ST&I measurements [41, 42] and relationship discovery [16, 43], and prediction models for technological forecasting [17, 44].

This study employed the following intelligent bibliometric models, in conjunction with descriptive statistics and co-occurrence analyses, which identify similar items (e.g., keywords and authors) by assessing their frequency of occurrence together.

#### *i. Hierarchical Topic Tree*

The Hierarchical Topic Tree (HTT) model is a network-based method of hierarchical community detection, designed to unveil research topics and their inherent hierarchical relationships within a corpus of research articles [16]. This model operates on a weighted co-term network as its input and identifies a collection of community anchors based on two distinct topological characteristics: 1) a notably high density and 2) a relatively greater distance with other high-density nodes. Each anchor subsequently forms the nucellus of a community, with the remaining nodes allocated to a community based on the shortest topological distance. This iterative process continues, generating sub-communities until no discernible anchors remain. The sub-communities established at each iteration contribute to the construction of topical layers, with the anchors serving as representative labels. Ultimately, the outcome is a hierarchically partitioned co-term network that reflects the layered intellectual structure within a field of knowledge.

[4] https://openalex.org/about

*ii. Scientific Evolutionary Pathways*

The innovation literature posits that scientific innovation is an accumulative process driven by the progression of knowledge [45]. The Scientific Evolutionary Pathways (SEP) model was designed to track this accumulative process by identifying a predecessor-descendent relationship between a newly created topic and previous ones [15]. The SEP model organizes articles with the same publication year into one time slice, treating the entire dataset as a bibliometric stream. Each topic is defined as a collection of articles, and the model analyses each article in the stream by classifying it into its most similar topic (i.e., predecessor). It determines whether the article contributes new knowledge by assessing its semantic distance from the centroid of the topic, and then groups articles identified as contributing new knowledge into several new topics, i.e., descendants.

Utilizing Gephi [46], the SEP model visualizes predecessor-descendent relationships in a network:

- A node represents a topic with its size proportional to the number of articles in the topic.
- An edge represents the predecessor-descendent relationship between its connected nodes. Its weight is determined by the similarity between the two topics.
- Nodes in the same color signify their belonging to the same community. The assignment of colors is based on a modularity-based community detection algorithm integrated into Gephi.

## IV. Bibliometric Portraits of Responsible AI

We constructed the bibliometric portraits of responsible AI from four aspects: Key technological players and their interactions with research communities and the responsibility principles, the topical hierarchy and dynamics of responsible AI and its historical dynamics, the interplay between primary AI techniques and the responsibility principles, and the profile of a core collection of the responsible AI research.

Given a bird's eye view, Fig. 1 illustrates the publication trends of the 17,799 articles related to responsible AI and contributed by the AI community since 2015. As anticipated, a discernible upward trend emerges starting in 2015 and accelerating after 2020. The incomplete data for 2023 would be an exclusion in this chart.

### *A. Key Technological Players and Interactions*

Defining key technological players as the most productive countries/regions, institutions, and research communities in a specific scientific or technological area. Fig. 2 paints a geographical portrait of the key players in the responsible AI research. At the forefront, China and the USA lead the game as dominant players, collectively contributing to over 40% of the total publications. India, the UK, and Germany follow, each contributing more than 1000 publications to the field. The 10 most productive countries collectively account for 86.5% of the research articles on responsible AI, aligning with the Pareto principle.

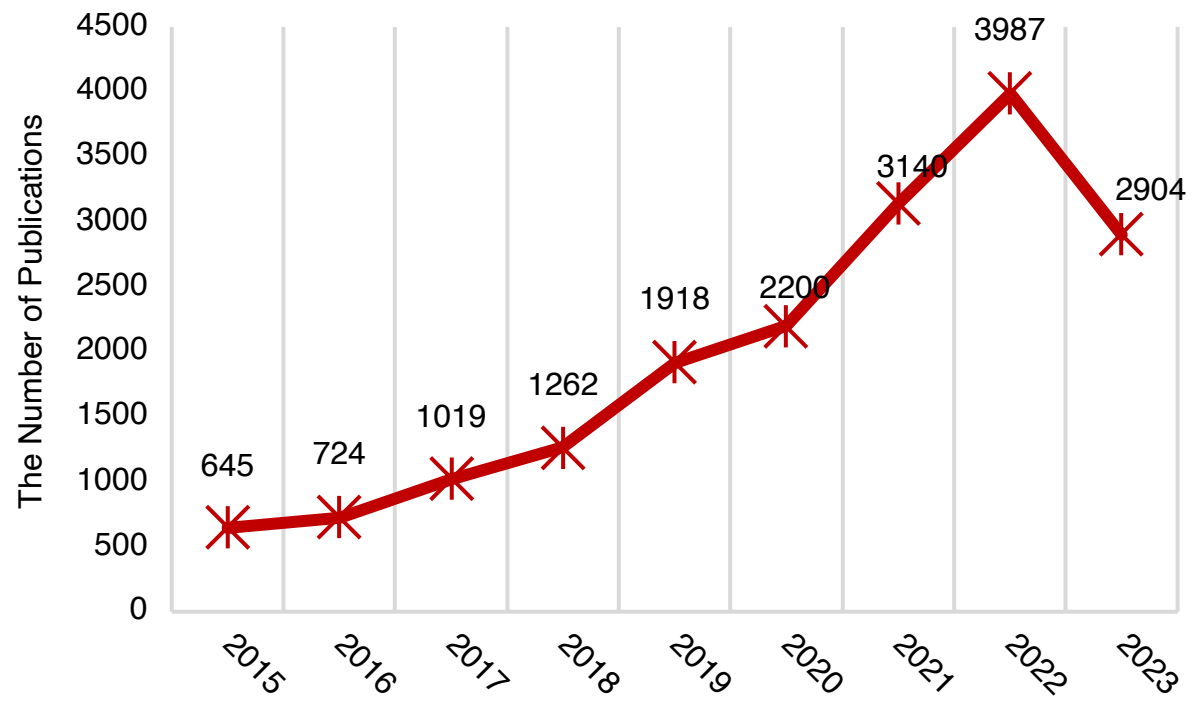

Fig. 1. Publication Trends of Responsible AI-related Articles since 2015.

Table II lists the top 15 most productive institutions in responsible AI research. In conjunction with Fig. 2, an observation is drawn regarding the prevalence of universities in contributing to the topic, but three national research institutions stand out: the French National Centre for Scientific Research (France), the Chinese Academy of Sciences (China), and the Commonwealth Scientific and Industrial Research Organization (CSIRO, Australia). This presence suggests potential national interests in responsible AI research. For example, as a possible response to China's Ministry of Science and Technology's release of the *Governance Principles for a New Generation of Artificial Intelligence: Develop Responsible Artificial Intelligence* in June 2019[5], Chinese universities have notably surged ahead in Table II.

TABLE II
Top 10 Most Productive Institutions in Responsible AI Research

| | ***Institutions*** | ***Countries*** | ***#P*** |
|---|---|---|---|
| 1 | French National Centre for Scientific Research | France | 183 |
| 2 | Xidian University | China | 156 |
| 3 | Chinese Academy of Sciences | China | 151 |
| 4 | Carnegie Mellon University | USA | 122 |
| 5 | University of Technology Sydney | Australia | 121 |
| 6 | University of Oxford | UK | 108 |
| 7 | Tsinghua University | China | 106 |
| 8 | Delft University of Technology | Netherland | 105 |
| 9 | University of Electronic Science and Technology of China | China | 105 |
| 10 | University of Science and Technology of China | China | 105 |

*Notes*. #P: The number of publications.

[5] See the details from the website of the US Library of Congress: https://www.loc.gov/item/global-legal-monitor/2019-09-09/china-ai-governance-principles-released/

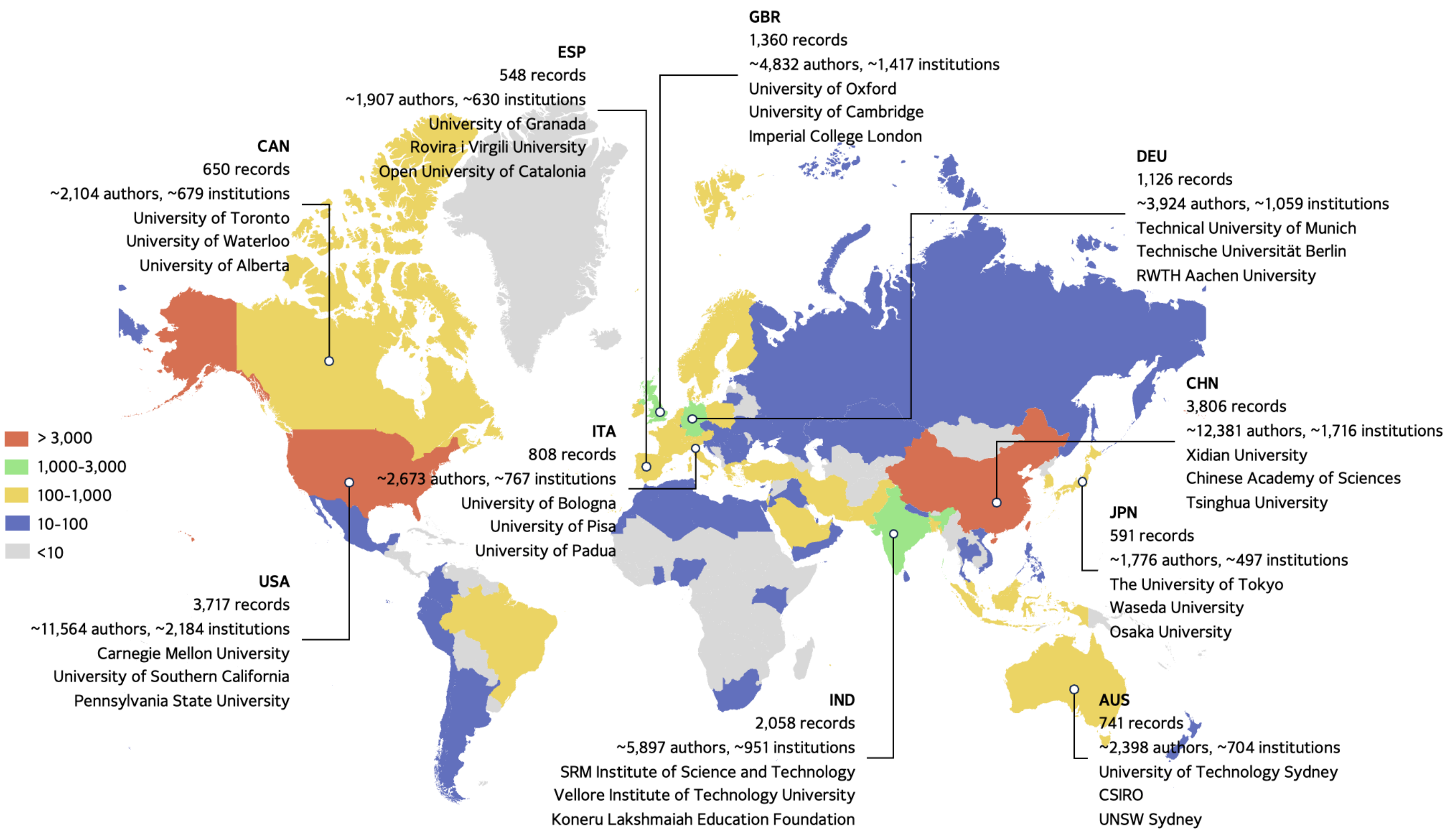

Fig. 2. Geographical Portrait of the Key Technological Players in Responsible AI Research.

Our investigation into research communities extends to the analysis of involved journals and conference proceedings to responsible AI. Table III presents the top 15 journals and conference proceedings, which have published the largest number of articles on responsible AI. Considering AI's subareas, there is a notable representation of journals spanning various facets of AI: *IEEE TNN* and *Artificial Intelligence* for theoretical developments in AI, particularly in machine learning studies; *Information Sciences* and *Knowledge-based Systems* for AI's technological innovation and system applications; and *IEEE TKDE* and *Pattern Recognition* for broad topics related to data mining and knowledge discovery. Despite our primary focus on the AI community, the inclusion of the journal *AI & Society* highlights the cross-disciplinary nature of responsible AI, showcasing active engagements with studies in humanities and social sciences.

In terms of conference proceedings, with their annual nature, the number of publications linked to one proceeding is comparable to as that of journals. However, a discernible surge in responsible AI research is observed after 2021, with the computer vision and communications societies emerging as the most active research communities for this new topic. This trend sheds light on the rising prominence of journals in machine learning and intelligent systems.

TABLE III
TOP 15 JOURNALS AND CONFERENCE PROCEEDINGS WITH THE LARGEST NUMBER OF RESPONSIBLE AI-RELATED PUBLICATIONS

| ***Journals*** | ***#Pub.*** | | ***Conferences*** | ***#Pub.*** |
|---|---|---|---|---|
| Information Sciences | 362 | 1 | 2021 IEEE BigData | 55 |
| AI & SOCIETY | 280 | 2 | 2023 IEEE/CVF WACV | 43 |
| Expert Systems with Applications | 234 | 3 | 2022 IEEE GLOBECOM | 40 |
| Knowledge Based Systems | 179 | 4 | 2021 IEEE GLOBECOM | 40 |
| Neurocomputing | 160 | 5 | 2022 IJCNN | 39 |
| IEEE Transactions on Knowledge and Data Engineering (TKDE) | 109 | 6 | 2022 ICASSP | 36 |
| Journal of Intelligent and Fuzzy Systems | 99 | 7 | 2021 IEEE TrustCom | 30 |
| Journal of Ambient Intelligence and Humanized Computing | 92 | 8 | 2022 IEEE/CVF CVPR | 30 |
| Soft Computing | 91 | 9 | 2021 IEEE SSCI | 26 |
| IEEE Transactions on Neural Networks and Learning Systems (TNN) | 87 | 10 | 2021 IEEE/CVF ICCV | 22 |
| Artificial Intelligence | 86 | 11 | 2022 ACM WSDM | 21 |
| Applied Intelligence | 84 | 12 | 2022 ACM/IEEE HRI | 20 |
| Pattern Recognition | 83 | 13 | 2022 GECCO | 20 |
| Frontiers in Artificial Intelligence | 80 | 14 | 2022 ACM MM | 19 |
| SN Computer Science | 76 | 15 | 2022 IEEE ICDE | 18 |

To portray the core research community of responsible AI, we linked journals (Fig. 3) and conference proceedings (Fig. 4) with a set of selected responsibility related FoS tags in OpenAlex. While the distribution of involved countries and publication venues mirrors that of the entire dataset, a distinct pattern emerges - most articles in this collection cluster around topics “privacy” and “security”, with only a few articles scattered across topics like “explainability”, “transparency”, and “trustworthiness”. This is consistent with the involvement of the communications society in the realm of cybersecurity and privacy, suggesting that these issues may have triggered the initial interaction between the AI community and real-world concerns related to responsibility. Simultaneously, this finding motivates us to delve deeper into the profiling of responsible AI through a series of topic analyses.

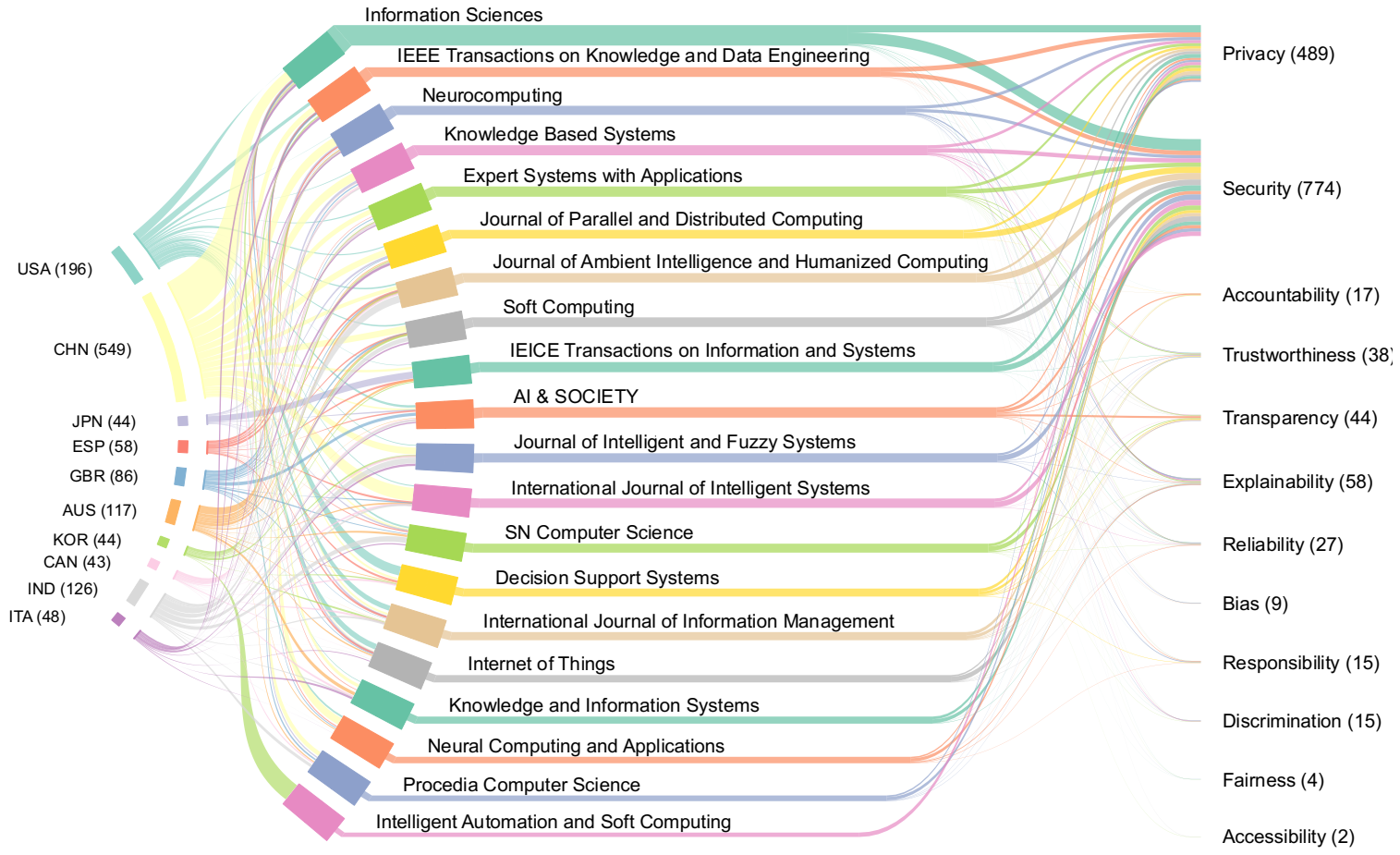

Fig. 3. Portraits of the Responsible AI Community with Journals.

## B. *Topical Hierarchy and Dynamics of Responsible AI*

Leveraging OpenAlex’s FoS tags to enable a pre-cleaned hierarchical topical system for detailed topic analyses, we utilized an article’s FoS tags as linked topics and constructed a FoS-based co-occurrence network for the application of the HTT and SEP models: HTT generates a data-driven topical hierarchy, offering insights into the knowledge structure of responsible AI research. SEP identifies predecessor-descendant relationships between research topics, and these evolutionary relationships provide understandings to the historical dynamics of responsible AI research, e.g., the origin of the field, the timeline of involved concepts and technologies, and the potential future directions it might evolve.

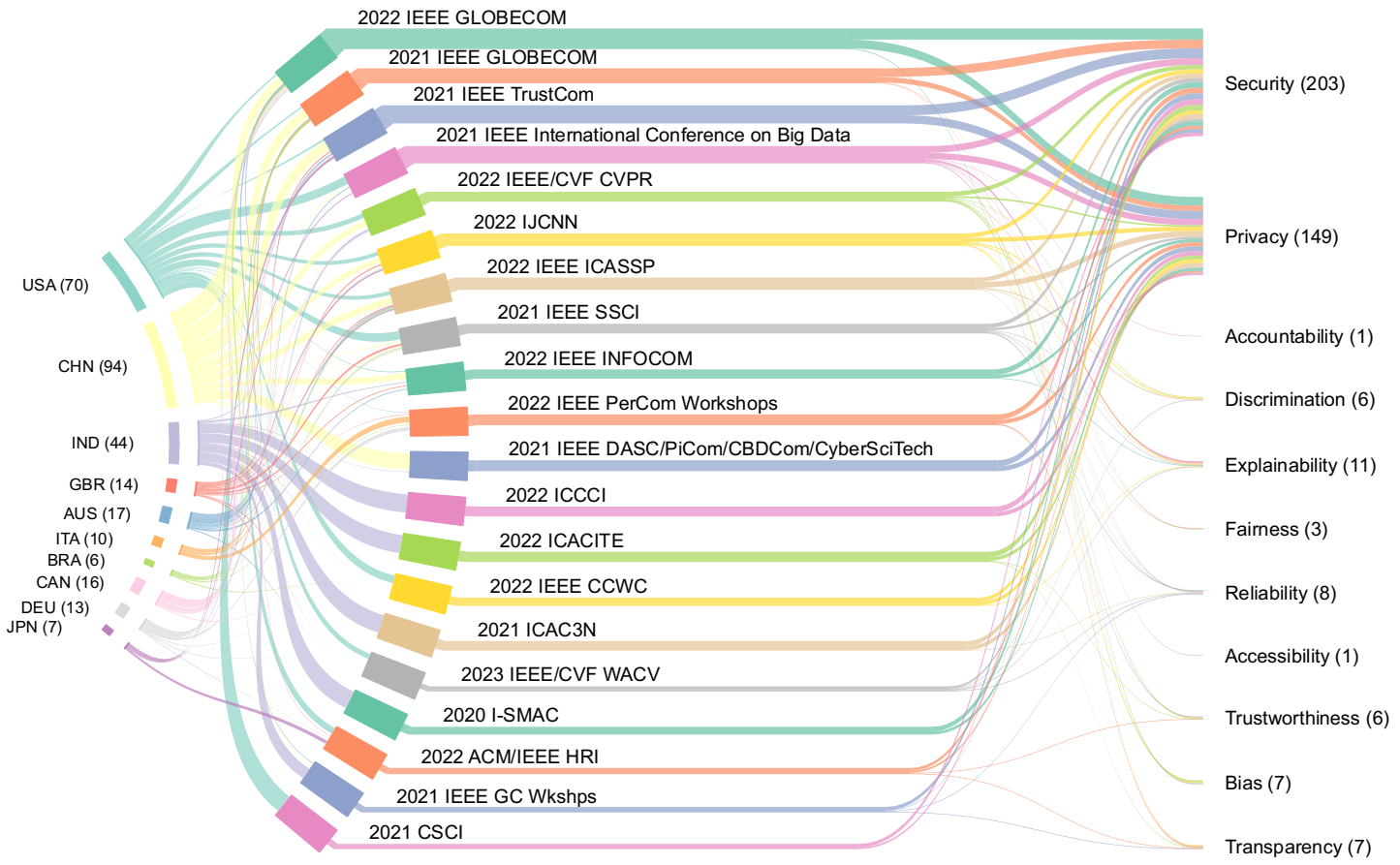

Fig. 4. Portraits of the Responsible AI Community with Conference Proceedings.

Fig. 5 shows the hierarchical topic tree of responsible AI research, delineating AI’s knowledge structure into four fundamental components: machine learning, mathematics, data mining, and computer networks. Note that HTT highlights the discovery of a hierarchical relationship between two research topics within a specific dataset, and thus, this hierarchy does not necessarily indicate a sense of a belongingness between the components.

We specifically annotated the following observations:

- *Machine Learning*: This branch draws a clear hierarchical structure of the machine learning research, consisting of neural networks[6], deep learning, and their downstream applications, e.g., human-computer interactions, bioinformatics, and recommender systems. Notably, we observed several strong connections with some responsibility issues, e.g., accountability when applying to knowledge management and specific tasks like recommender systems, interpretability for inference-related studies, and trustworthiness when performing human-computer interactions.
- *Data Mining*: This branch is application-focused, highlighting the broad applications of AI and data science techniques in real-world cases, e.g., database management, information retrieval, cyber-physical systems, and governance. Information privacy is a key concern, closely linked with internet privacy, information sensitivity and confidentiality, information leakage, privacy software, and personal identities. In addition, interoperability is involved with database and data modelling.

[6] The topic of support vector machines (SVM) under neural networks can be explained as, after 2015, SVM, together with techniques in its leaf nodes (e.g., random forests, Bayesian, and genetic algorithms), frequently co-occurs with neural networks, as comparative baselines, for example.

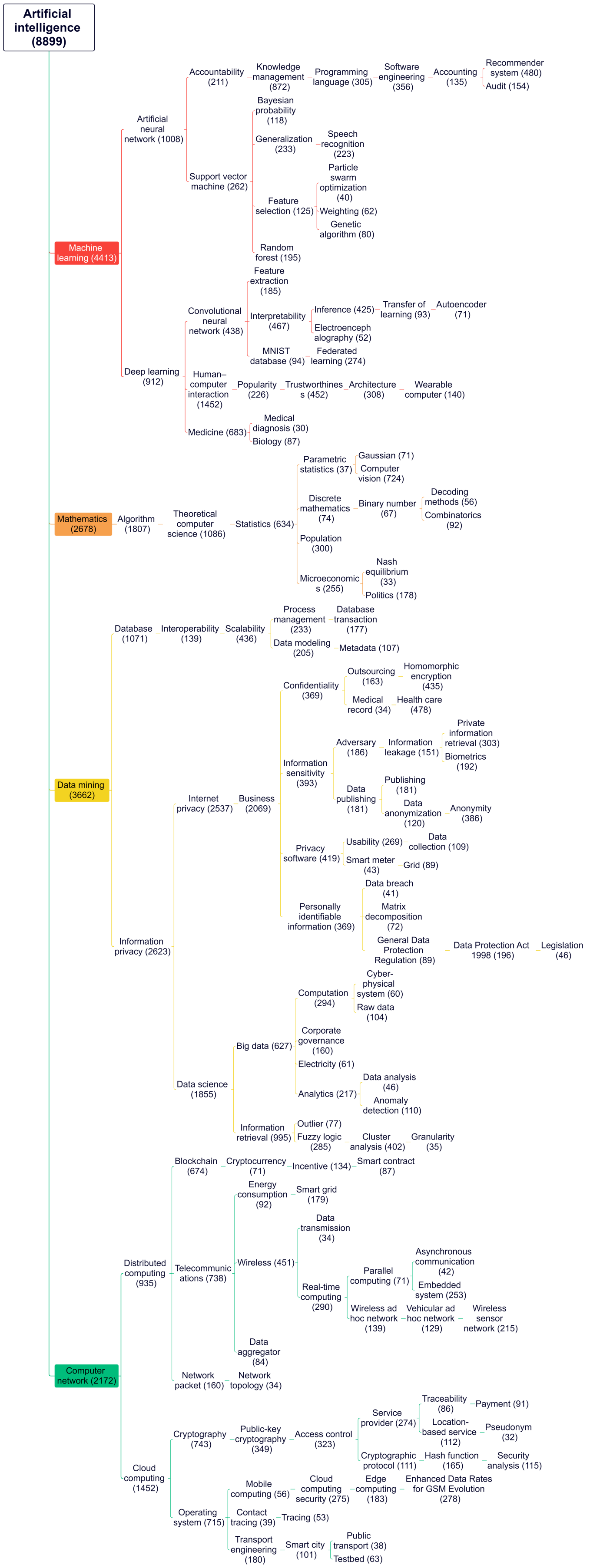

Fig. 5. Hierarchical Topic Tree of Responsible AI Research.

- *Computer Networks*: This branch groups topics in distributed systems, cloud computing, and telecommunications, which is the key contributor to privacy and security related studies in software engineering and electrical engineering.
- *Mathematics*: This branch highlights the theoretical base of computer science and contains topics such as statistics, parametric statistics, and Gaussian.

Fig. 6 illustrates the evolutionary pathways of responsible AI research spanning the years 2015 to 2023. We employed a K-means clustering algorithm on articles published in 2015. Utilizing the elbow method, we identified five topics – artificial intelligence, computer security, internet privacy, encryption, and computer network – as the starting points of the evolutionary pathways. These topics serve as a technical profile of responsible AI before its formal terminology and concepts were established. These initially distinct technologies have gradually coalesced into their respective technological cohorts since 2015, forming the foundational clusters that construct the technological landscape of responsible AI.

We discussed the following observations:

- *Machine Learning and Statistics* (pink cluster): This cluster reveals the evolution of machine learning-based algorithms, techniques, and applications from 2015 to 2023. Key sub-technologies and their evolutionary pathways encompass neural networks, human-computer interaction, robots, and various AI algorithms. Notably, interpretability has emerged as one of the most critical concerns within the context of neural networks.
- *Data and Information Privacy* (blue cluster): This cluster highlights the evolution of privacy-related topics and their interactions, e.g., the mechanisms of privacy protection in blockchains, privacy safeguards in various data and information sources (e.g., the web, social media, and healthcare records), and some regulations of data protection. Interoperability, trustworthiness, and accessibility are intertwined throughout these pathways.
- *System Security* (green cluster): This cluster progresses from the privacy concerns of data architecture to the security design and implementation of analytical systems, interacting with data science, internet of things, telecommunications, and cloud computing techniques. It emphasizes the development of methods, software, and hardware for privacy and security protection. Some notably techniques include federated learning, adversarial system, private information retrieval, etc.
- *Encryption* (yellow cluster): While relatively small, this cluster draws a clear pathway of cryptography, including both its fundamental techniques and applications.
- *Ethics* (orange cluster): This cluster predominantly addresses AI ethics from the perspective of humanities and social sciences, spanning areas such as business, management science, politics, and law.

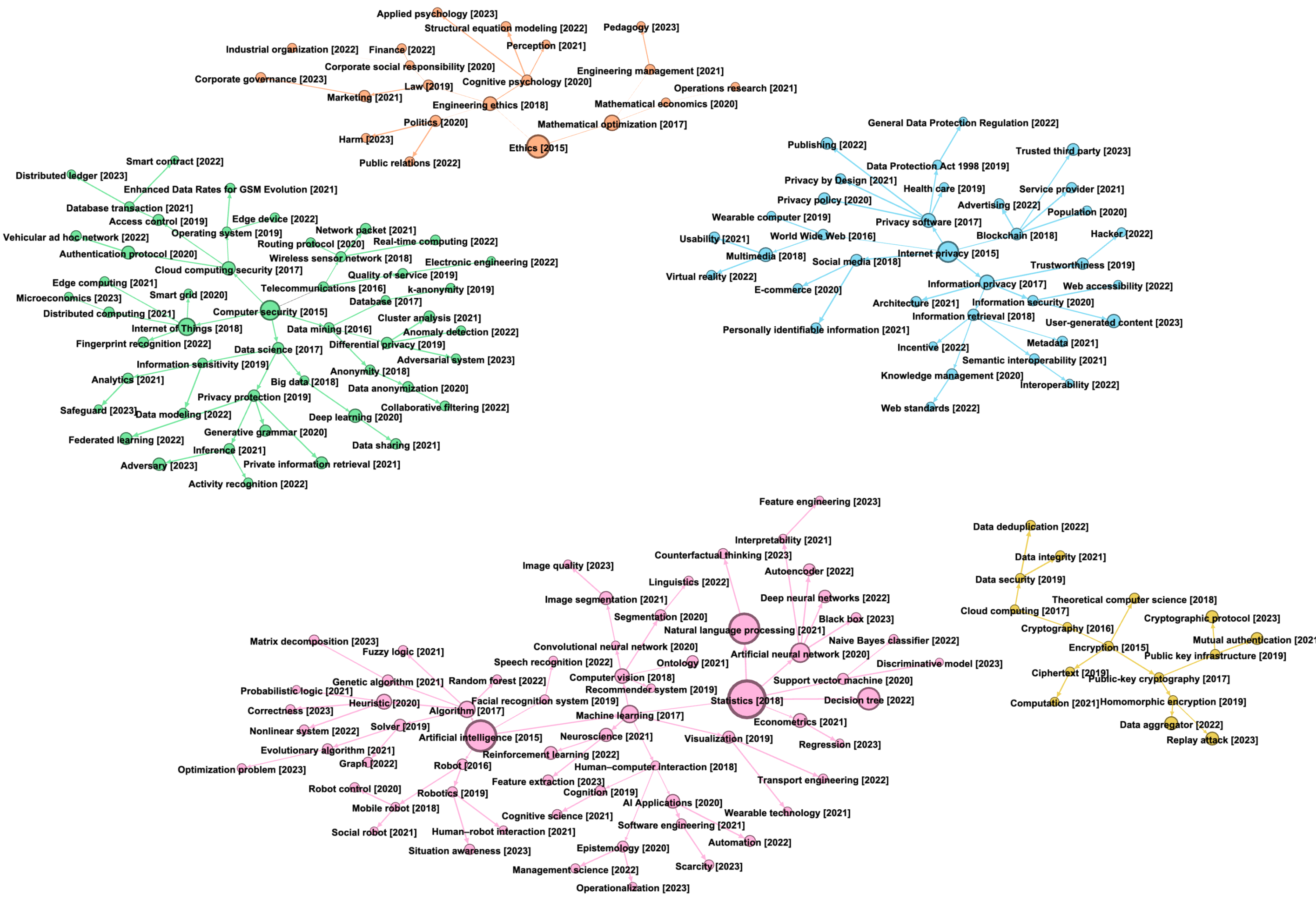

Fig. 6. Scientific Evolutionary Pathways of Responsible AI Research (2015 - 2023).
Notes. The year associated with a node shows the time when the node was born, indicating when the related knowledge initially became influential within the entire dataset.

### C. *Responsibility Principles vs. Primary AI Techniques*

Revisiting the responsibility principles defined in Section II-B and aligning them with the FoS tags, we identified 12 principles for mapping. We added "responsibility", but excluded "intelligibility" due to its extremely low frequency and "inclusiveness" since it mainly aligns with social science topics when existing computer science literature typically addresses similar concepts under the term "fairness" [47]. Then, we selected 16 primary AI techniques based on their frequency in our dataset. To visualize the co-occurrence between the responsibility principles and primary AI techniques, we utilized the Circos approach [48], as illustrated in Fig. 7.

We derived the following key observations:

- As depicted in Fig. 5 and Fig. 6, the initial studies on responsible AI were primarily centered around privacy and security. Techniques related to cloud computing, machine learning, internet of things, and blockchains contribute more than 50% of related research.
- Machine learning, including its sub-techniques such as neural networks, deep learning, and reinforcement learning, has directly engaged with all principles but demonstrated a tangible connection with explainability (86%), discrimination (83%), bias (74%), and fairness (52%). Clearly, this is consistent with the community's continuous interests on explainable AI and ethical AI.

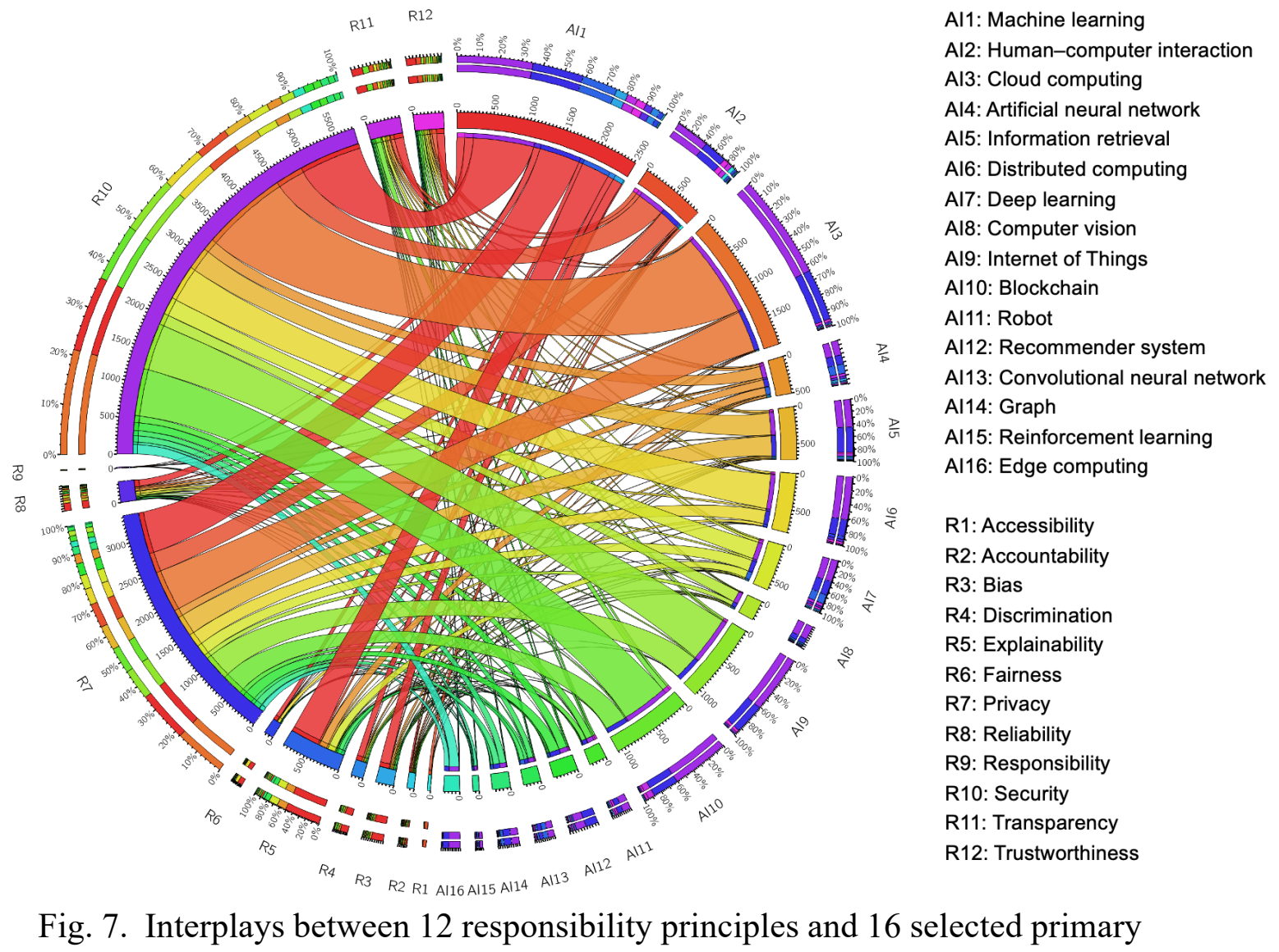

Fig. 7. Interplays between 12 responsibility principles and 16 selected primary AI techniques.
Note. The statistics is in Supplementary Table 1.

- Human-computer interaction occupies the accessibility principle (83%) and is also the second largest contributor to trustworthiness and transparency.
- Some significant matching pairs also include cloud computing with accountability, distributed computing with fairness, and blockchain with accountability and transparency.
- Interestingly, the responsibility principle is linked with machine learning (31%), robot (31%), and blockchain (15%), indicating some of the AI backbones that initiate the topic of responsible AI.

### *D. Core Cohort of Responsible AI Research*

While our study primarily relied on the dataset of responsible AI-related articles contributed by the AI community since 2015, we maintained a keen interest in exploring the core concept of responsible AI and its developmental progress. Therefore, with a specific focus on the WoS Core Collection to ensure the high standards of coverage, on November 23, 2023, we curated 380 articles directly employing the term "responsible artificial intelligence" or "responsible AI" in their titles, abstracts, and keywords.

Fig. 8 draws a series of portraits for the core cohort, illustrating its publication trend (Fig. 8A), article types (Fig. 8B), and WoS subject categories (Fig. 8C). Intriguingly, the term "responsible artificial intelligence" was firstly mentioned in 2015 [49] [7], but as evidenced in Fig. 8A, academia did not commence investigations into responsible AI until 2019 and a significant upswing started in 2021. Compared to the large volume of research articles on AI, responsible AI remains a relatively recent and evolving topic.

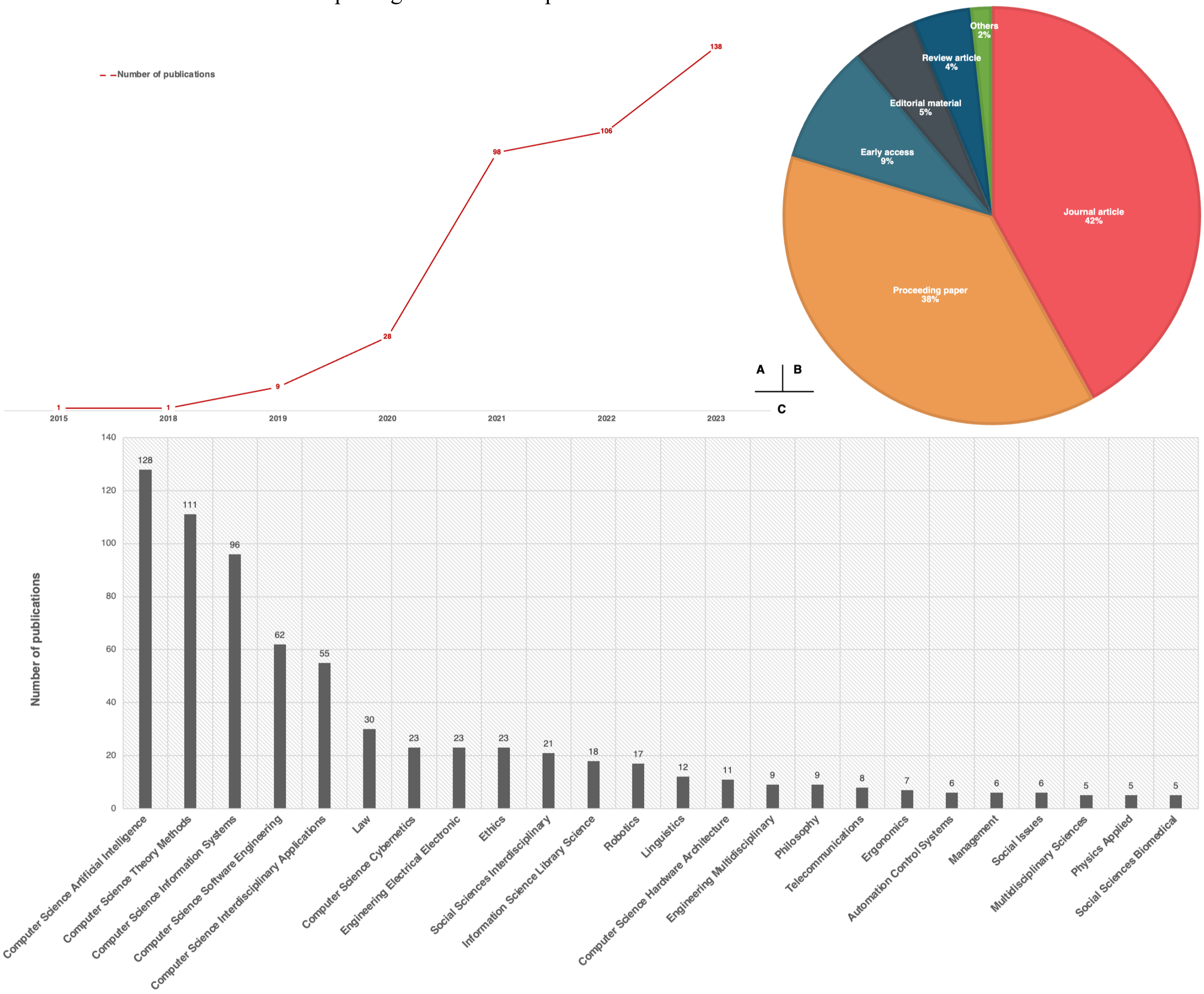

Fig. 8. Portraits of the Core Cohort of Responsible AI Research: A) Publication Trend, B) Distribution of Article Types; and C) Distribution of the WoS Subject Categories (with at least 5 articles).
Notes. 1) An article may have more than one article type, and thus, Fig. 8B exhibit overlaps. 2) While the core cohort spans across 89 WoS subject categories, the 24 subject categories depicted in Fig 7C encapsulate 334 articles – 87.9% of the core cohort.

[7] We conducted a similar search on Google Scholar and no results were retrieved for articles predating the year 2015.

Certainly, the computer science community is the primary contributor to the core cohort. The category "computer science, artificial intelligence" leads the ranking, comprising 128 articles (33.7%), and the six sub-categories within computer science collectively encompass 275 articles, accounting for 72.4% of the cohort. Remarkably, the inclusion of categories such as law, ethics, social sciences, and information sciences underscores the cross-disciplinary nature of responsible AI research.

Table IV presents the top 15 most productive institutions in responsible AI research within the core cohort. The USA takes the lead, followed by the UK, Australia, and the Netherlands. Interestingly, China is conspicuously absent from the list. Given that the core cohort reflects the most recent conceptualization of responsible AI rather than its fundamental technologies and sub-technologies, the dominance of the West, supported by their international-renowned universities, has demonstrated their capabilities in shaping innovative theories and concepts and driving global technological advancements. In addition, Microsoft emerges as one of the most productive institutions, uniquely representing the corporate sector and signaling its ambition in this cutting-edge domain.

TABLE IV
TOP 15 MOST PRODUCTIVE INSTITUTIONS ON RESPONSIBLE AI RESEARCH IN THE CORE COHORT

| | *Affiliations* | *Country* | *#Pub.* |
|---|---|---|---|
| 1 | The University of Texas Austin | USA | 16 |
| 2 | CSIRO | Australia | 13 |
| 3 | Umea University | Sweden | 11 |
| 4 | Northeastern University | USA | 10 |
| 5 | University of Oxford | UK | 9 |
| 6 | Microsoft | USA | 8 |
| | University of Washington | USA | |
| 8 | Norwegian University of Science and Technology | Norway | 7 |
| | University of Bologna | Italy | |
| | University of Amsterdam | Netherlands | |
| 11 | University of New South Wales | Australia | 6 |
| | Cornell University | USA | |
| | Stanford University | USA | |
| | University of Sheffield | UK | |
| | Delft University of Technology | Netherlands | |
| | University of Cambridge | UK | |

Aiming to understand the content of the core cohort, we employed the KeyBERT model [50] to automatically extract 1257 terms from the titles and abstracts of the 380 articles, and created a co-term network using VoSViewer [51], see Fig. 9. Slightly different from previous portraits that feature a large amount of AI-related technical terms, Fig. 9 unveils AI's extensive engagements across multiple disciplines and research areas, e.g., medicine and healthcare (with bioethics and CoVID-19), governance (with industry, economy, law, policy, and sustainability), etc. This, again, highlights the cross-disciplinary nature of responsible AI, transitioning from fundamental AI techniques and algorithms to a high-level perspective encompassing the design, development, implementation, and regulation of AI in close coordination with academia, industry, and society. Particularly intriguing is the relatively scattered knowledge landscape in Fig. 9, without clearly defined boundaries and communities. This aligns with the theory of complex systems and their transitions, indicating the chaos inherent in system disruption when new knowledge and technologies emerge [18, 52].

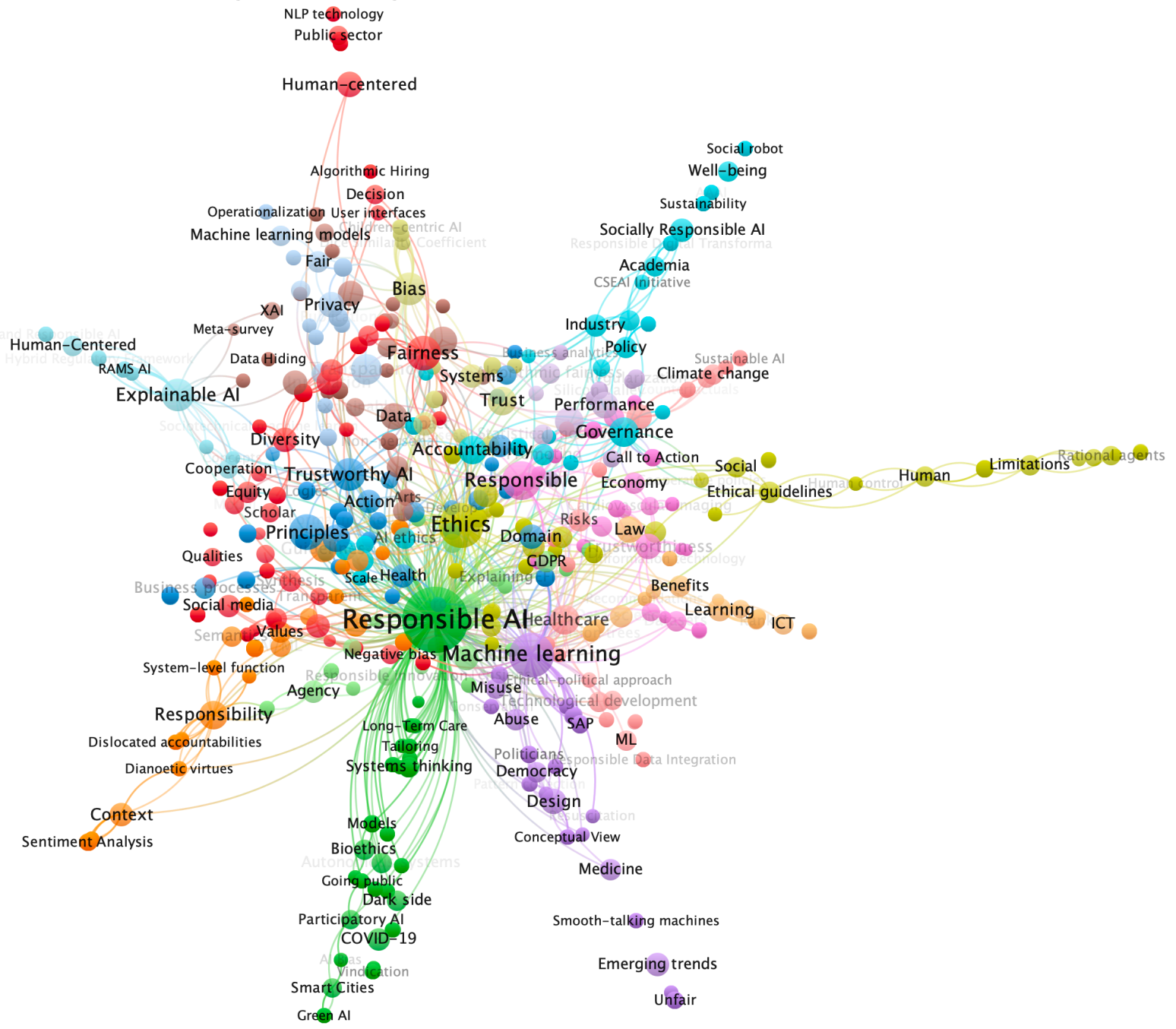

Fig. 9. Co-term Network of Responsible AI Research in the Core Cohort.
Notes. A node represents a term and the edge between two nodes indicates their co-occurrence relationships. The size of a node corresponds to the frequency of its related term, while its color refers to potential semantic similarities measured by the strength of co-occurrence.

## V. DISCUSSION AND CONCLUSIONS

To unveil the technological landscape of responsible AI and particularly the endeavors undertaken by the AI community, this study developed an intelligent bibliometrics-based analytical framework including descriptive statistics, co-occurrence analyses, a model of hierarchical topic tree, and scientific evolutionary pathways. This study recognized key technological players and their interactions, identified the topical hierarchy and evolution of responsible AI, discovered the interplays between the responsibility principles and primary AI techniques, and profiled a core cohort of the most recent cross-disciplinary developments on responsible AI.

### *A. Intersection between Responsible AI and related Concepts*

Aligned with the conceptual discussion on responsible AI and its interplay with related concepts such as trustworthy AI, explainable AI, and ethical AI, the bibliometric portraits derived from the primary dataset (e.g., Figs. 3-7) reveal a significant emphasis on trustworthy AI, particularly regarding security and privacy considerations. Concurrently, explainable AI indicates profound engagements with machine learning and related techniques, whereas ethical AI appears somewhat detached from AI's broader societal aspects and applications. Intriguingly, upon closer examination of the core cohort of responsible AI-related studies, Fig. 9 illustrates a cohesive synthesis of these AI concepts emerges within the framework of responsible AI, and this synthesis extends to encompass conceptual principles, AI techniques, and societal applications.

### B. Technical and Practical Implications

Traditional bibliometric studies, relying on descriptive statistics for knowledge domain profiling, have been widely applied. However, they often fall short in uncovering latent patterns and mechanisms. While topic models offer a semantic understanding of a data corpus, they come with challenges related to human interventions in term pre-processing and parameter settings. Moreover, high model accuracy does not necessarily translate into promising results for real-world applications. In this study, the proposed analytical framework with intelligent bibliometrics incorporates AI and data science techniques, and these models are either nonparametric or possess limited parameters. More importantly, these models have been independently evaluated in various training datasets [15, 16] and empirical cases [53, 54], and their consistent results in our study contribute to draw a comprehensive narrative of responsible AI.

This application brings some fresh ideas to traditional bibliometric studies: 1) To leverage AI's analytical capabilities into bibliometrics for enhanced knowledge discovery, e.g., extending co-occurrent relationships to intricate relationships, e.g., hierarchy and evolution. 2) The cross-validation with experimental comparisons and empirical examinations would ensure the practical feasibility of a computational model in addressing real-world issues and further add values by uncovering insights behind the results derived from data analytics.

### C. Limitations and Future Directions

Responsible AI is still a new topic, with the nature of high dynamics, radical development, and active cross-disciplinary interactions. We acknowledged the limitations of this study and identified the following future directions. 1) Future studies should focus on monitoring the ongoing accumulation of the core cohort (Section IV-D) and profiling its inter-/multi-/cross-/trans-disciplinary interactions by elaborating extra data sources (e.g., political documents and social media), along with multimodal data (e.g., images from full-text research articles and videos from social media). 2) The incredible capabilities of large language models could further enhance the analytical framework, e.g., accurate data annotation, entity extraction, and knowledge summarization.

## Acknowledgment

The authors acknowledge the use of ChatGPT for polishing the language of this paper. The authors clarify that all aspects of this work were designed, developed, implemented, and reported by the authors, without any contributions from AI-generated content.

## Supplemental Materials

Supplementary Table 1 and the high-solution versions of Figs 2, and 5-9 can be retrieved through: https://github.com/IntelligentBibliometrics/Responsible-AI-Review

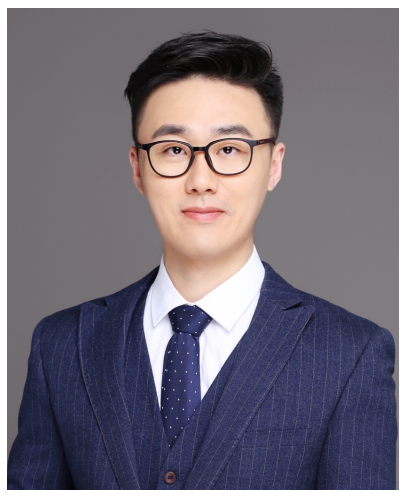

**Yi Zhang** (SM'23) is an Associate Professor at the Australian Artificial Intelligence Institute, University of Technology Sydney (UTS). He received a dual-PhD degree in Management Science and Engineering from Beijing Institute of Technology (2016) and Software Engineering from UTS (2017). His research aligns with bibliometrics and technology management. He has published more than 100 research articles in leading journals and conferences in related fields.

Dr Zhang was the receipt of the 2019 Australian Research Council's Discovery Early Career Researcher Award (DECRA). He serves as the Specialty Chief Editor of *Frontiers in Research Metrics and Analytics*, and an Associate Editor of *Technological Forecasting and Social Change*, *IEEE Transactions on Engineering Management*, and *Scientometrics*.

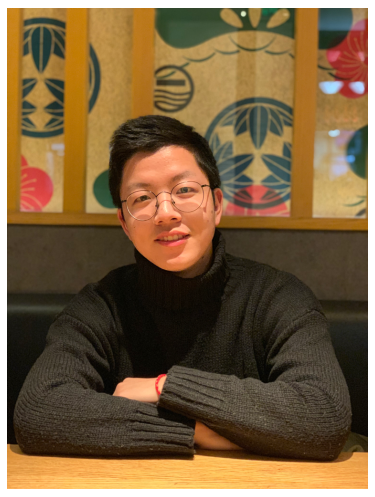

**Mengjia Wu** received the B.Sc. and MA degrees in information science from Huazhong University of Science and Technology, Wuhan, China, and he completed his Ph.D. study at the Australian Artificial Intelligence Institute, University of Technology Sydney.

Dr Wu has published more than 20 conference/journal papers in bibliometric and cross-disciplinary venues. His research interest is leveraging bibliometrics, text analytics, network analytics, and graph neural networks to develop and optimize knowledge extraction and discovery models. In 2021, he was granted the ISSI student travel prize from the International Society for Informetrics and Scientometrics.

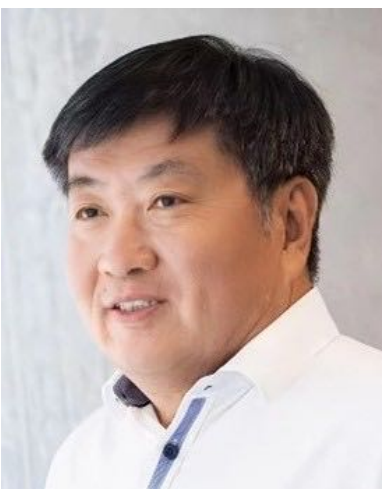

**Guangquan Zhang** is an Associate Professor and Director of the Decision Systems and e-Service Intelligent Research Laboratory at the University of Technology Sydney, Australia. He received the Ph.D. degree in applied mathematics from Curtin University of Technology, Australia, in 2001. His research interests include fuzzy machine learning, fuzzy optimization, and machine learning. He has authored five monographs, five textbooks, and 460 papers including 220 refereed international journal papers.

Dr Zhang has won seven Australian Research Council (ARC) Discovery Projects grants and many other research grants. He was awarded an ARC QEII fellowship in 2005.

He has served as a member of the editorial boards of several international journals, as a guest editor of eight special issues for IEEE transactions and other international journals and co-chaired several international conferences and workshops in fuzzy decision-making and knowledge engineering.

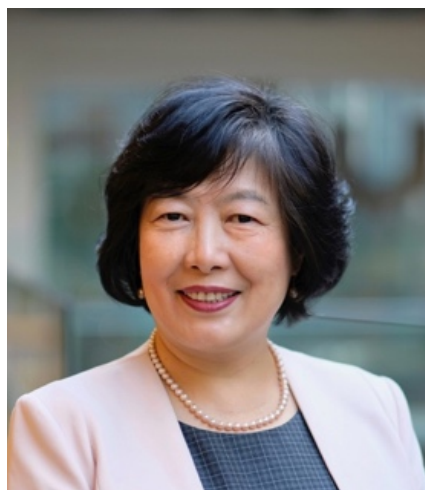

**Jie Lu** (F'18) is an Australian Laureate Fellow, IFSA Fellow, ACS Fellow, Distinguished Professor, and the Director of Australian Artificial Intelligence Institute at the University of Technology Sydney, Australia. She received a PhD degree from Curtin University in 2000. Her main research expertise is in transfer learning, concept drift, fuzzy systems, decision support systems and recommender systems. She has published over 500 papers in IEEE Transactions and other leading journals and conferences. She is the recipient of two IEEE Transactions on Fuzzy Systems Outstanding Paper Awards (2019 and 2022), NeurIPS2022 Outstanding Paper Award, Australia's Most Innovative Engineer Award (2019), Australasian Artificial Intelligence Distinguished Research Contribution Award (2022), Australian NSW Premier's Prize on Excellence in Engineering or Information & Communication Technology (2023), and the Officer of the Order of Australia (AO) 2023.